%% file: paper.tex
\documentclass[runningheads]{llncs}
\usepackage{graphicx}

\usepackage{tikz}
\usepackage{comment}
\usepackage{amsmath,amssymb} 
\usepackage{color}

\usepackage[accsupp]{axessibility}  

\input{preamble}

\begin{document}
\pagestyle{headings}
\mainmatter
\def\ECCVSubNumber{1766}  

\title{L3: Accelerator-Friendly Lossless Image Format for High-Resolution, High-Throughput DNN Training}

\titlerunning{L3: Accelerator-Friendly Lossless Image Format for DNN Training}
%
\author{Jonghyun Bae \and
Woohyeon Baek \and
Tae Jun Ham \and
Jae W. Lee}
\authorrunning{J. Bae et al.}
%
\institute{Seoul National University, Seoul, Korea \\
\email{\{jonghbae,\;baneling100\}@snu.ac.kr,\;ham.taejun@gmail.com,\;jaewlee@snu.ac.kr}
}

\maketitle

\begin{abstract}
The training process of deep neural networks (DNNs) is usually pipelined with stages for data preparation on CPUs followed by gradient computation on accelerators like GPUs. In an ideal pipeline, the end-to-end training throughput is eventually limited by the throughput of the accelerator, not by that of data preparation. In the past, the DNN training pipeline achieved a near-optimal throughput by utilizing datasets encoded with a lightweight, lossy image format like JPEG.  However, as high-resolution, \emph{losslessly}-encoded datasets become more popular for applications requiring high accuracy, a performance problem arises in the data preparation stage due to low-throughput image decoding on the CPU. Thus, we propose L3, a custom \textbf{\underline{l}}ightweight, \textbf{\underline{l}}oss\textbf{\underline{l}}ess image format for high-resolution, high-throughput DNN training. The decoding process of L3 is effectively parallelized on the accelerator, thus minimizing CPU intervention for data preparation during DNN training. L3 achieves a 9.29$\times$ higher data preparation throughput than PNG, the most popular \emph{lossless} image format, for the Cityscapes dataset on NVIDIA A100 GPU, which leads to 1.71$\times$ higher end-to-end training throughput. Compared to JPEG and WebP, two popular \emph{lossy} image formats, L3 provides up to 1.77$\times$ and 2.87$\times$ higher end-to-end training throughput for ImageNet, respectively, at equivalent metric performance.
\keywords{DNN training, Data preparation, Image processing}
\end{abstract}

\input{contents/1-intro}
\input{contents/2-related}

\input{contents/3-background}

\input{contents/4-design}

\input{contents/5-eval}

\input{contents/6-conclusion}

\clearpage
%
%
\bibliographystyle{splncs04}
\bibliography{paper}
\end{document}

%% file: preamble.tex
\usepackage{xspace}
\usepackage{multirow}
\usepackage[hyphens]{url}

\usepackage{cite}

\newcommand{\para}[1]{\noindent \textbf{#1.}}

\newcommand{\name}{L3\xspace}

\newcommand{\ballnumber}[1]{\tikz[baseline=(myanchor.base)]\node[circle,fill=.,inner sep=1pt] (myanchor) {\color{-.}\bfseries\footnotesize #1};}

\usetikzlibrary{tikzmark}
\tikzset{mycircled/.style={circle,draw,inner sep=1pt,line width=0.04em}}
\newcommand{\ballwhite}[1]{\tikzmarknode[mycircled,draw=black]{t1}{\bfseries\footnotesize #1}}



%% file: contents/1-intro.tex
\section{Introduction}
\label{sec:l3:intro}

The recent development of deep neural networks (DNNs) has been incited by the large-scale, publicly accessible datasets such as ImageNet~\cite{imagenet2009jia}, CIFAR~\cite{cifar}, and SVHN~\cite{svhn}. These datasets are conventionally encoded using a lossy encoding format (e.g., JPEG). However, for emerging application domains requiring high accuracies, such as autonomous driving~\cite{Chen_2017_CVPR,kitti, Menze_2015_CVPR, Zhou_2018_CVPR}, image generation and restoration~\cite{esrgan, edsr, swinir, uformer}, denoising~\cite{eunet, deam}, and medical diagnosis~\cite{hou2019high, brain2020, medi2020}, the use of lossy image format can potentially result in accuracy loss. Thus, the demands for \emph{losslessly} encoded datasets continue to increase for more accurate pixel-wise segmentation and object representations.

Today's end-to-end DNN training pipeline is composed of the data preparation stage on the CPU, followed by the gradient computation stage on the accelerator (e.g., GPU, TPU). Since the next mini-batch data preparation stage overlaps with the current batch's gradient computation stage, the longest stage limits end-to-end training throughput. In the traditional setting of DNN training pipeline employing lossy-encoded datasets, the gradient computation stage is the primary bottleneck. Thus, most of the research has been conducted to improve the model training throughput on the accelerator by reducing inter-node communication overhead~\cite{gpip32019huang, pipedream2019deepak, dlt2019}, optimizing GPU memory~\cite{dragon2018markthub, vdnn2016rhu, super2018wang}, and compiler optimization for DNN operators~\cite{tvm2018chen, rammer2020ma}.

However, data preparation has recently become the critical performance bottleneck, especially for datasets with high-resolution, lossless images, which require a complex decoding process. According to a recent study~\cite{trainbox2020park}, the throughput ratio of the gradient computation to the data preparation can be as high as 54.9$\times$ even with well-optimized data preparation. Balancing the pipeline stages, in this case, would require 50$\times$ more CPU cores. There are several recent proposals to address stalls in the data preparation, such as NVIDIA's Data Loading Library (DALI)~\cite{dali:web} and TrainBox~\cite{trainbox2020park}. However, they only target lossy-encoded datasets with custom hardware support using a dedicated hardware JPEG decoder on NVIDIA A100 GPU~\cite{a100hwacc} and an FPGA accelerator~\cite{trainbox2020park}.

Thus, we propose \name, a new \textbf{\underline{l}}ightweight, \textbf{\underline{l}}oss\textbf{\underline{l}}ess image format, whose decoding can be fully accelerated on data-parallel architectures such as GPUs. \name eliminates the data preparation stalls by replacing the complex decoding algorithm running on the CPU to achieve the end-to-end training throughput close to the ideal pipeline with zero overhead for data preparation. For standalone execution \name on NVIDIA A100 GPU delivers a 9.29$\times$ higher decoding throughput for the Full HD (1920$\times$1080) Cityscapes dataset than the lossless PNG format on Intel Xeon CPU while maintaining at most 9\% loss in the compression ratio. Furthermore, \name achieves a 1.25$\times$ (1.77$\times$) and 1.93$\times$ (2.87$\times$) higher geomean (maximum) training throughput than JPEG and lossy WebP, respectively, for seven state-of-the-art object detection and semantic segmentation models at equivalent metric performance.

%% file: contents/2-related.tex
\section{Related Work}
\label{sec:related}

\para{Optimized Data Preparation} There exist several recent proposals on improving the data preparation performance for DNN training. DIESEL~\cite{diesel2020wang} splits files into two parts---in-memory metadata and on-disk objects, to fully utilize the I/O bandwidth by reducing redundant metadata accesses. CoorDL~\cite{mohan2020analyzing} uses host memory as a cache for the storage and utilizes a new caching policy named MinIO, which does not replace the once-cached dataset elements. \emph{tf.data}~\cite{tfdata2021murray} proposes an optimized input data pipeline for TensorFlow with optimized parallel I/O, caching, and automatic resource (e.g., CPU, I/O) management to minimize the data preparation latency. These proposals reduce the latency of \texttt{Load} but do not address the \texttt{Decode} which can be the main bottleneck, especially when the input images employ lossless encoding. In contrast, \name presents a new lossless, accelerator-friendly image format to provide superior \texttt{Decode} throughput, hence effectively eliminating the bottleneck in the \texttt{Decode} stage.

\para{Optimized Lossless Image Decoding} Many proposals attempt to improve the decoding throughput of the lossless image format, such as Huffman decoding~\cite{cano2021, massive2018, huffman2020yamamoto}, bzip2~\cite{bzipaccel}, and LZSS~\cite{culzss2011} by utilizing GPUs or specialized hardware. However, there are trade-offs between decoding throughput and compression ratio. So, achieving both high decoding throughput and high compression ratio simultaneously is challenging. Instead, Gompresso~\cite{gompresso2016Sitaridi} utilizes a slightly modified LZ77 algorithm and partition-based Huffman coding to improve the compression/decompression throughput on GPUs. Similarly, Adaptive Loss-Less (ALL) data compression~\cite{all2017shunji} exploits run-length and adaptive dictionary-based coding for GPUs, as well as a partition-based compression/decompression scheme to improve throughput. However, both research target general-purpose applications; hence, they are sub-optimal for DNN training. Instead, \name is a more specialized format for DNN training to achieve high throughput while maintaining a competitive compression ratio.

%% file: contents/3-background.tex
\section{Background and Motivation}
\label{sec:l3:back}

\begin{figure}[t]
    \centering
    \includegraphics[width=0.94\textwidth]{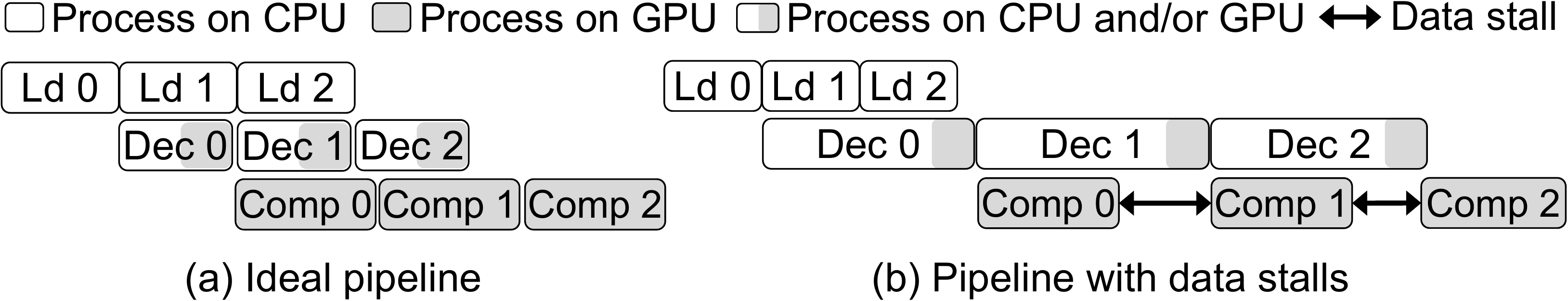}
    \caption{(a) Ideal DNN training pipeline (b) DNN training Pipeline with data stalls (Ld: \texttt{Load}, Dec: \texttt{Decode}, Comp: \texttt{Compute})}
    \label{fig:pipeline}
\end{figure}

\subsection{DNN Training Pipeline}
\label{sec:l3:back:pipe}

The DNN training operates as a three-stage pipeline: (1) \texttt{Load} image, (2) \texttt{Decode} image, and (3) \texttt{Compute} gradients. For each iteration, a single batch of images are loaded from a storage device to the host main memory (\texttt{Load}), and then those images are decoded with CPU and/or GPU~\cite{nvjpeg:web,nvjpeg2k:web} (\texttt{Decode}). These decoded images are transferred to the GPU device, and then a gradient computing iteration is performed (\texttt{Compute}). In most DNN frameworks such as PyTorch~\cite{pytorch:web}, TensorFlow~\cite{tf2016martin}, and MXNet~\cite{mxnet2015chen}, these three steps execute in a pipelined manner.

Since those three stages are pipelined, the overall training throughput is determined by the stage that takes the longest time among \texttt{Load}, \texttt{Decode}, and \texttt{Compute}. Ideally, the \texttt{Load} and \texttt{Decode} time should be shorter than \texttt{Compute} time so that the time spent on \texttt{Load} and \texttt{Decode} is completely hidden, as shown in Figure~\ref{fig:pipeline}(a). However, as shown in Figure~\ref{fig:pipeline}(b), either \texttt{Decode} or \texttt{Load} stage may take longer to bottleneck the pipeline. Such cases are likely to happen when high-resolution images are decoded on CPU.

\begin{figure}[t]
    \centering
    \includegraphics[width=0.94\textwidth]{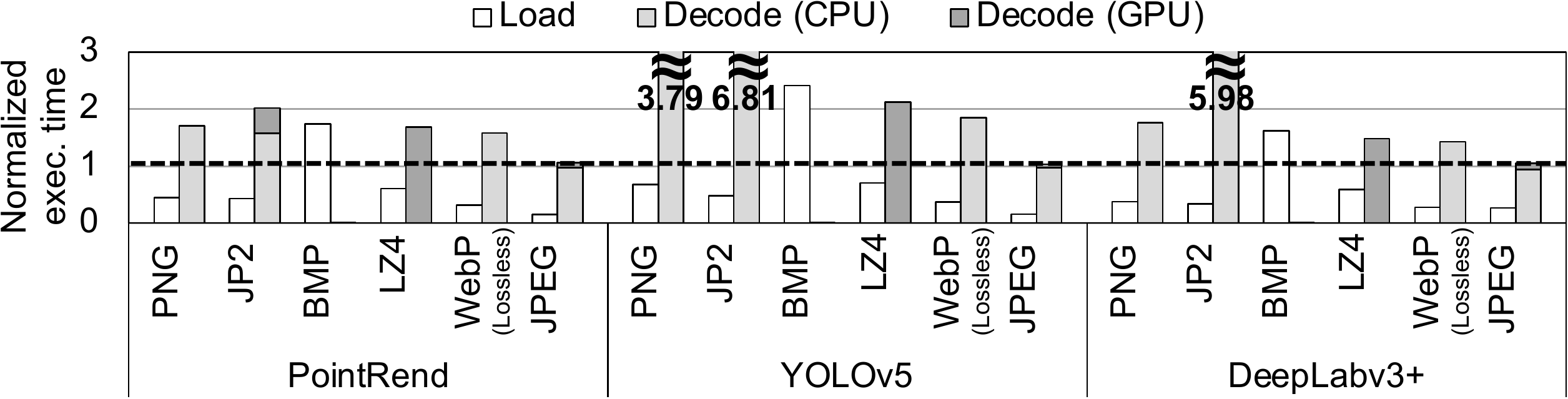}
    \caption{\texttt{Load} and \texttt{Decode} execution time normalized to the \texttt{Compute} time (in the dotted line) for three representative models in Table~\ref{tab:batch} and six image formats. }
    \label{fig:break}
\end{figure}

\subsection{Data Preparation Bottleneck}
\label{sec:l3:back:prep}

\para{Profiling Data Preparation Bottleneck} Figure~\ref{fig:break} compares the \texttt{Load} and \texttt{Decode} time with the \texttt{Compute} time for two segmentation models (PointRend~\cite{pointrend2020Kirillov}, DeepLabv3+~\cite{deeplab2018chen}) and one detection model (YOLOv5~\cite{yolov5:web}). The models run on a platform with Intel Xeon Platinum 8275CL CPU, NVIDIA A100 GPU with 40GB HBM2 DRAM, and 1TB NVMe SSD. The datasets and mini-batch size used for each model are described in Section~\ref{sec:l3:eval:metho}. The graph shows that the \texttt{Load} and \texttt{Decode} time varies greatly across input image formats, even for the same model. For example, the PNG, JP2, and lossless WebP spend more time on \texttt{Decode} than \texttt{Compute}; this implies that the DNN training pipeline is bottlenecked by \texttt{Decode}. On the other hand, the BMP has a longer \texttt{Load} time than \texttt{Compute} time, indicating that the pipeline is bottlenecked by \texttt{Load}. This is because the uncompressed BMP data tends to be much larger than other compressed image formats. Among those measured, the JPEG is the only one that is not bottlenecked in data preparation (i.e., \texttt{Load} and \texttt{Decode}). However, JPEG is a lossy image format that loses some information in the original data.

\para{Decoding Bottleneck for Lossless Image Formats} JPEG format requires much less time to decode as it utilizes GPU to perform most of the decoding process. Since modern high-end GPUs have substantially higher computation power than CPUs, this greatly helps to prevent \texttt{Decode} from being a bottleneck. However, much less attention has been paid to accelerating \emph{lossless} image formats. NVIDIA proposes the use of LZ4-based format with its library to accelerate the decoding of LZ4-compressed images using GPU~\cite{nvcomp:web}; however, as shown in Figure~\ref{fig:break}, its decoding speed is much slower than that of JPEG. Also, it is much more challenging to accelerate \texttt{Decode} of other lossless image formats. For example, it is well-known that PNG decoding, especially a process of decoding data compressed with dynamic Huffman coding, is difficult to parallelize and thus not well suited for GPU implementation~\cite{all2017shunji, gompresso2016Sitaridi, huffman2020yamamoto}.

\begin{table}[t]
    \centering
    \caption{Test set accuracy and its standard deviation for seven object detection and semantic segmentation models on PNG (Lossless), JPEG (Lossy), and WebP (Lossy) encoded datasets}
    \label{tab:pngjpegacc}
    \begin{tabular}{l||c||c|c}
        \hline
        \multicolumn{1}{c||}{\multirow{2}{*}{\textbf{Model}}} & \textbf{Lossless} & \multicolumn{2}{c}{\textbf{Lossy}} \\ \cline{2-4} 
        \multicolumn{1}{c||}{} & \textbf{PNG (Stdev.)} & \textbf{JPEG (Stdev.)} & \textbf{WebP (Stdev.)} \\ \hline \hline
        DDRNet23-slim (mIoU)    & 0.729 (0.001) & 0.693 (0.003) & 0.689 (0.001)\\ \hline
        DeepLabv3+ (mIoU)       & 0.801 (0.004) & 0.801 (0.007) & 0.808 (0.009)\\ \hline
        MaskFormer (mIoU)       & 0.685 (0.008) & 0.669 (0.004) & 0.651 (0.007)\\ \hline
        PointRend (mIoU)        & 0.725 (0.001) & 0.711 (0.003) & 0.712 (0.002)\\ \hline
        EgoNet (PCK@0.1)        & 0.323 (0.005) & 0.300 (0.002) & 0.305 (0.002)\\ \hline
        PointPillar (mAP)       & 0.611 (0.006) & 0.593 (0.004) & 0.582 (0.009)\\ \hline
        YOLOv5 (mAP@50)         & 0.633 (0.004) & 0.605 (0.003) & 0.614 (0.007)\\ \hline
    \end{tabular}
\end{table}

\subsection{Comparison of Lossless and Lossy Image Formats} 
\label{sec:l3:back:acc}

Image formats can be lossy or lossless. A lossless image format (e.g., PNG, BMP, JPEG2000, lossless WebP) keeps all information in the raw image. In contrast, lossy image formats (e.g., JPEG, lossy WebP) often require less storage space but lose some information in the raw image. In the context of DNNs, high-resolution, lossless image formats are commonly utilized for domains where model accuracy is critical such as autonomous driving~\cite{Chen_2017_CVPR, kitti, Menze_2015_CVPR, Zhou_2018_CVPR}, image generation and restoration~\cite{esrgan, edsr, swinir, uformer}, denoising~\cite{eunet, deam}, and medical diagnosis~\cite{hou2019high, brain2020, medi2020}. Table~\ref{tab:pngjpegacc} shows the impact of lossy image format on the test set accuracy and its standard deviation of seven object detection and semantic segmentation models, which are often deployed for autonomous driving. We use the default JPEG and lossy WebP quality factor of the Python Pillow package, which is 75. The reported accuracy number is an average of three training runs, and we observe negligible variations in the accuracy. Other than the image format, we use the same hyperparameters taken from the original publication without tuning for both lossless and lossy image formats. The use of the JPEG and lossy WebP results in degradation of the test set accuracy in all models except DeepLabv3+, compared to a case where the lossless PNG format is used for encoding.

%% file: contents/4-design.tex
\section{\name Design}
\label{sec:l3:design}

\subsection{Design Goal}
\name is a new image format specialized for a specific use case (i.e., ML/DL training) with a lightweight, lossless encoding/decoding algorithm. Our design goal is to eliminate the data preparation bottleneck by (i) maximizing decoding throughput by leveraging the GPU while (ii) providing a good-enough compression ratio not to introduce a new bottleneck in the \texttt{Load} stage. The conventional lossless image formats feature different tradeoffs. If the compression ratio is most important, WebP would be the best option. If compatibility is most important, PNG would be the one. Instead, we carefully design \name to efficiently accelerate its decoding process on the GPU, unlike existing lossless image formats that are not suitable for GPU acceleration due to their limited parallelism. With \name, the training system can eliminate the CPU bottleneck to improve the DNN training throughput for datasets with lossless, high-resolution images.

\subsection{Encoding/Decoding Algorithm}
\label{sec:l3:design:algo}

\begin{figure*}[t]
    \centering
    \includegraphics[width=0.94\textwidth,page=1]{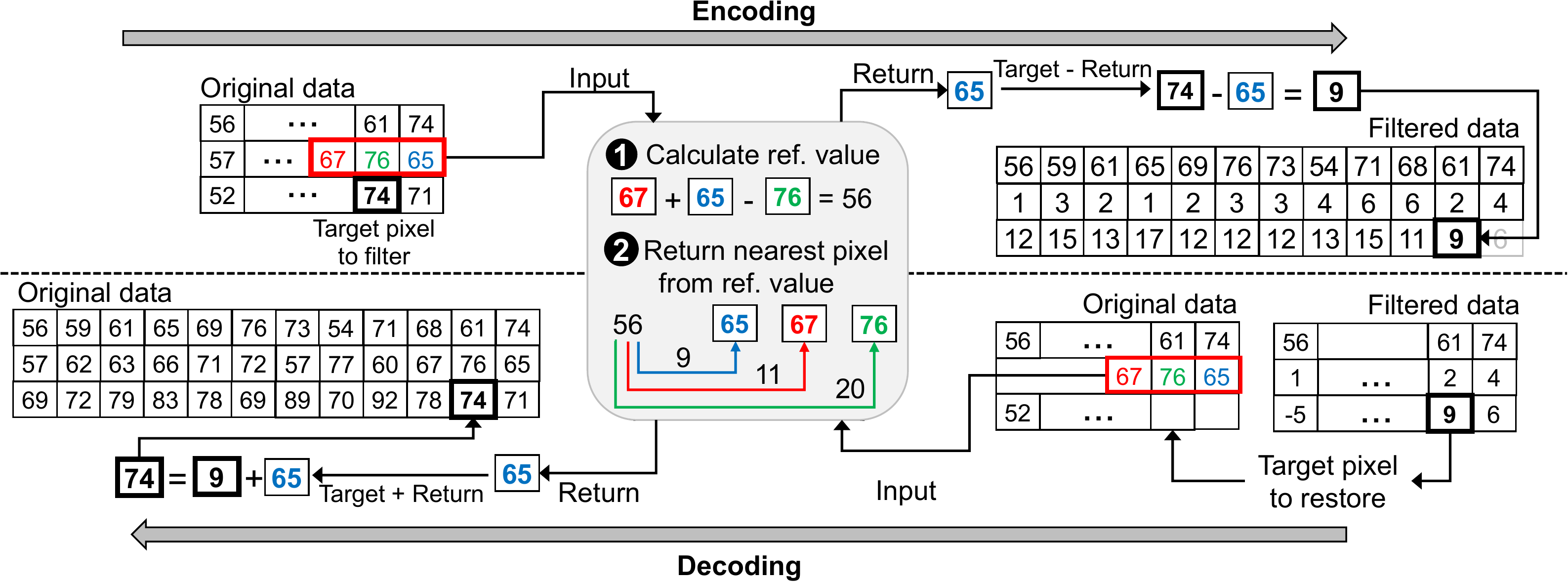}
    \caption{Example of the custom Paeth filter in \name}
    \label{fig:paeth}
\end{figure*}

Encoding and decoding for \name is a two-stage process: Paeth filtering followed by the row-wise base-delta encoding/decoding. Essentially, \name utilizes a customized Paeth filter to significantly reduce the number of bits required to represent a pixel, exploiting the spatial redundancy (i.e., nearby pixels have similar values). After Paeth filtering, \name performs base-delta encoding/decoding to reduce further the number of bits representing each pixel and uses a packed representation to store the image in a compressed format. The compressed image can be decoded by performing the reverse operation of each stage. In what follows, we describe each stage in greater detail.

\para{Custom Paeth Filter} The Paeth filter~\cite{paeth1991}, best known for its usage in PNG encoding, is a popular method to reduce the range of values for each pixel. The original Paeth filter encodes each pixel based on three neighboring pixels (left, top, top-left). However, this selection creates both \emph{row-wise} dependency along the row dimension (left) as well as \emph{column-wise} dependency (top) along the column dimension, thus making it challenging to exploit fine-grained (i.e., pixel-level) parallelism. Therefore, \name customizes the Paeth filter to make it more amenable to parallel execution by inspecting a different set of neighboring pixels (top-left, top, top-right), thus eliminating column-wise data dependency.

Figure~\ref{fig:paeth} shows the encoding/decoding process of the custom Paeth filter in \name. In the rest of this paper, we refer to this as the Paeth filter for brevity. First, the Paeth filter calculates the \emph{reference value} using those three neighboring pixels as follows: $Top\_Left + Top\_Right - Top$ (Step \ballnumber{1}). Then, among the three neighboring pixels, the filter selects the one whose value is the closest to the reference value (Step \ballnumber{2}). Finally, the difference between the original pixel value (i.e., 74 in the example) and the selected neighboring pixel value (i.e., 65 in the example) is stored. As an exception, the first row does not have a preceding row, so this row's filter operation is skipped.

The decoding process is similar to the encoding process. Since the first row is stored in a raw data format, we can start decoding from the second row towards the bottom row. The reference value can be computed by inspecting the three neighboring pixels in the preceding row to identify the pixel whose value is the closest to the reference. Then we add the stored residual (i.e., 9 in the example) to the pixel value (i.e., 65 in the example) to recover the original pixel value.

\begin{figure*}[t]
    \centering
    \includegraphics[width=0.94\textwidth,page=2]{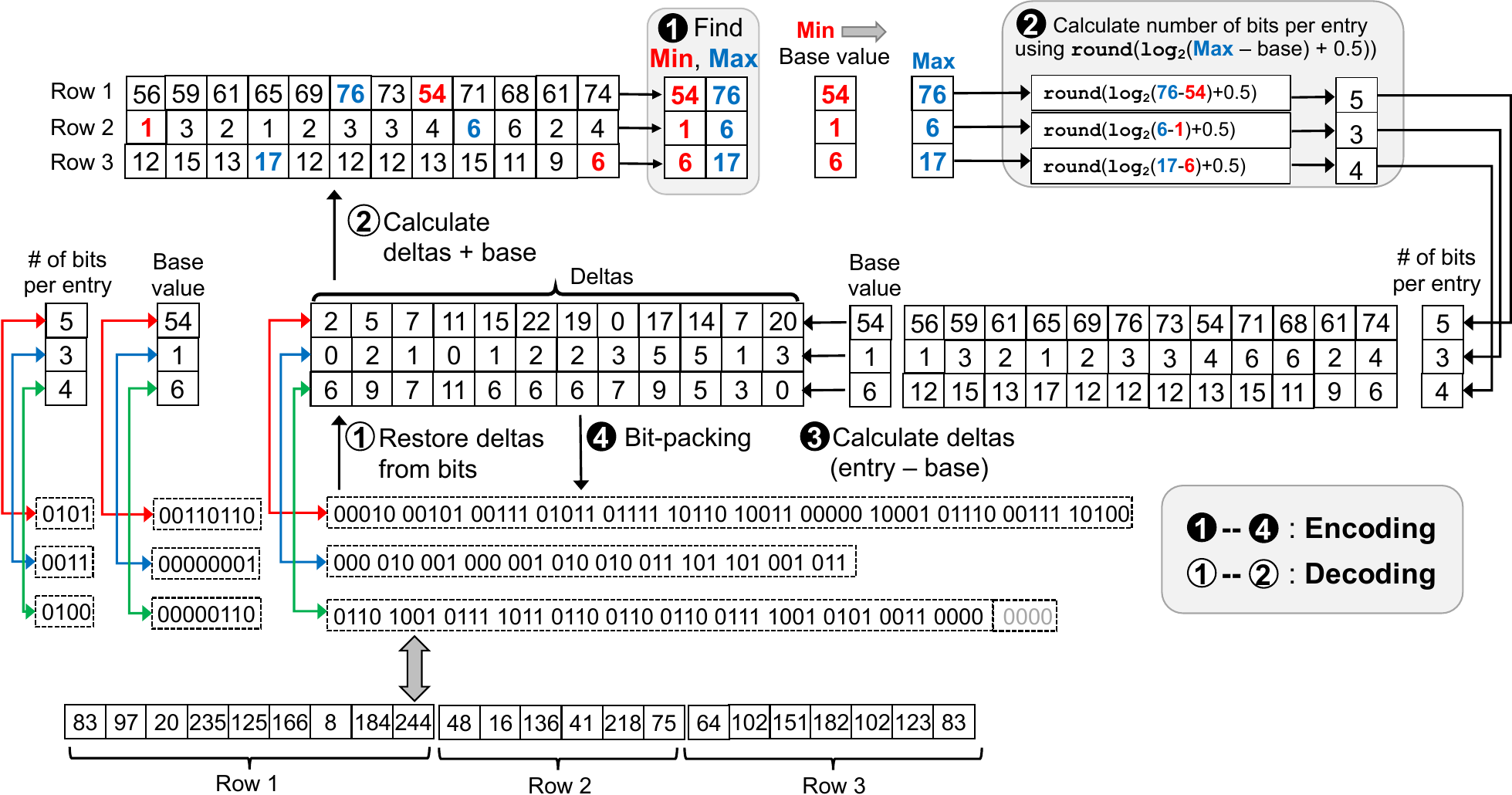}
    \caption{Example of base-delta encoding/decoding}
    \label{fig:bd}
\end{figure*}

\para{Base-Delta Encoding/Decoding} The outcome of the custom Paeth filter has a reduced value range. \name applies the base-delta encoding~\cite{bd1991farrens,bdi2012gennady} to each row to reduce the number of bits representing each pixel. Figure~\ref{fig:bd} illustrates this base-delta encoding and decoding process. Encoding is a four-step process. First, each row's minimum and maximum values are found (Step \ballnumber{1}). The minimum value is selected as the \emph{base value} of each row. Also, the minimum number of bits required to cover all delta values is computed (Step \ballnumber{2}). Then, the deltas from the base value for all elements are calculated (Step \ballnumber{3}). Finally, a compressed bitstream is generated by representing each delta using the minimum number of bits (Step \ballnumber{4}). Specifically, the first four bits represent the number of bits per delta for each row, and the following eight bits the base value. Then, the deltas of the row, each represented using the minimum number of bits, are appended. All the rows are concatenated to generate the final encoded stream, as shown at the bottom of Figure~\ref{fig:bd}.

The decoding process is quite simple. From the compressed stream, the decoder first reads the four bits as well as the following eight bits to identify the number of bits per entry and the base value for the row (Step \ballwhite{1}). From this point, the decoder extracts the delta one by one and adds the base value to reconstruct the original value (Step \ballwhite{2}). Once the first row is completed, the same process is repeated for the rest of the rows.

\begin{figure*}[t]
    \centering
    \includegraphics[width=0.94\textwidth,page=3]{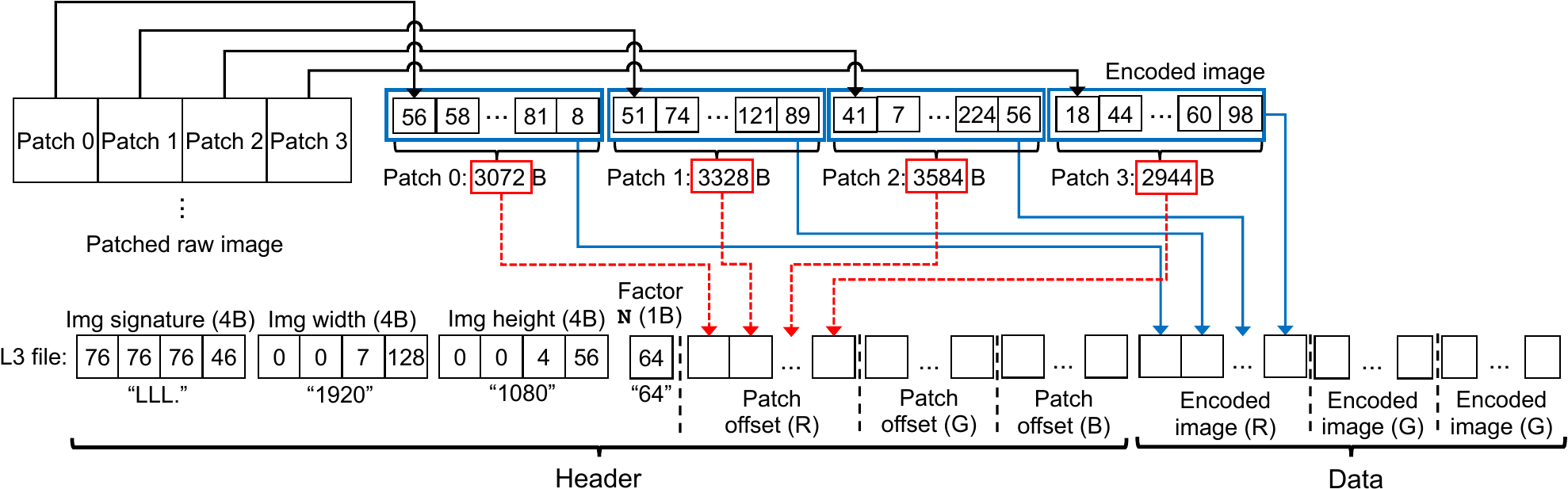}
    \caption{Example of \name file formatting}
    \label{fig:file}
\end{figure*}

\subsection{L3 File Format}
\label{sec:l3:design:file}

The \name encoding/decoding algorithm can work with any large-sized image. In practice, to maximize the parallelism and improve the decoding throughput in GPUs, \name divides a large image into multiple square \emph{patches} of size $N \times N$, where $N$ is the user-specified parameter. We empirically set $N$=32 for images whose resolution is below 1080$\times$720 (HD), $N$=64 for images whose resolution is between 1080$\times$720 (HD) and 1920$\times$1080 (FHD), and $N$=128 for images whose resolution is between 1920$\times$1080 (FHD) and 3840$\times$2160 (UHD). Once $N$ is set, the image is first separated into three channels (R, G, and B). Then, the image is divided into square-sized patches for each channel, and the encoding algorithm is applied to each patch. The encoded patch is concatenated, and the offset for each patch is recorded and stored in a separate array.

Figure~\ref{fig:file} shows the \name file format. The file consists of the header and the data sections. The header section contains a 4-byte file type signature (also called magic bytes), 4-byte image width, 4-byte image height, 1-byte patch size, and three patch offset arrays corresponding to the three color channels. The data section contains encoded patch data for patches.

\subsection{Optimizing L3 Decoder on GPU} 
\label{sec:l3:design:gpu}

\begin{figure*}[t]
    \centering
    \includegraphics[width=0.94\textwidth]{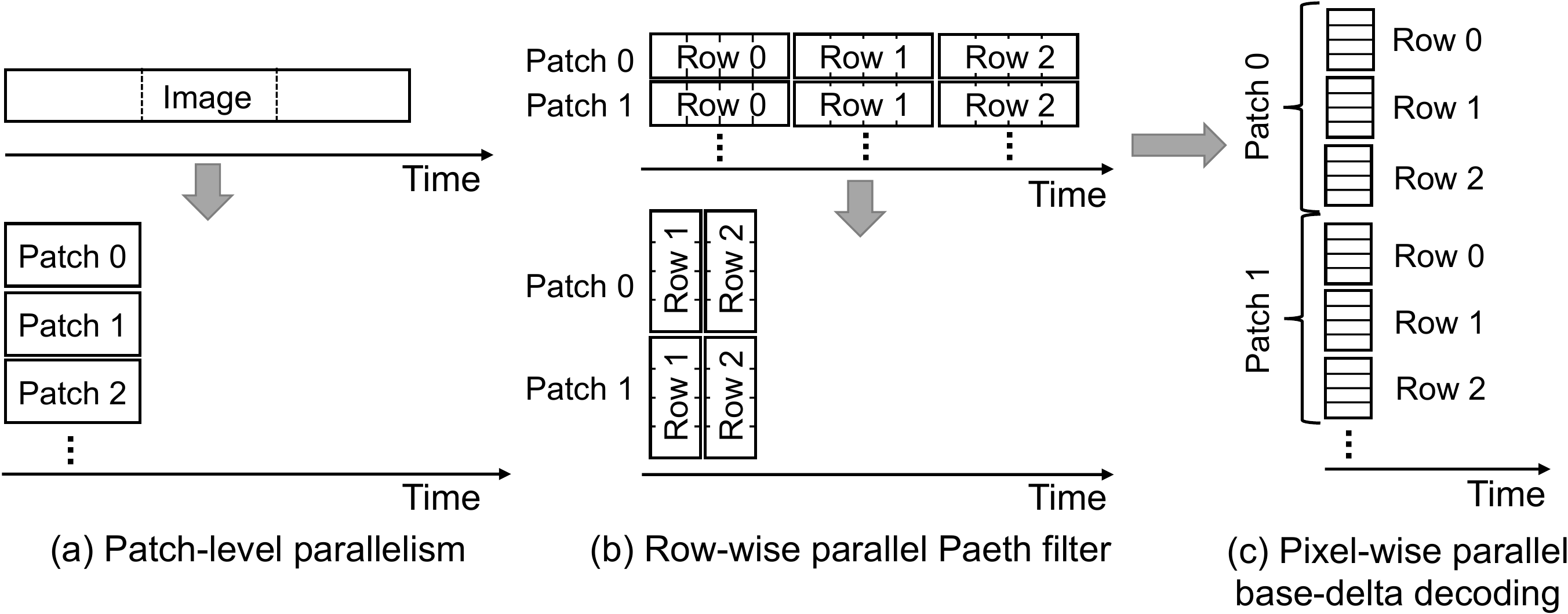}
    \caption{Parallel execution example of (a) Patch-level parallelism (b) Row-wise parallel Paeth filter (c) Pixel-wise parallel base-delta decoding.}
    \label{fig:parallel}
\end{figure*}

\para{Patch-level Parallelism} Figure~\ref{fig:parallel}(a) illustrates patch-level parallelism in \name. Without parallelism, a single thread block, a unit of scheduling for GPU devices, decodes the entire image over multiple iterations, underutilizing GPU resources. Instead, the \name decoder first reads the header of each image and then splits it into multiple patches based on the header. Then, the task of decoding a single patch is assigned to a distinct thread block. This enables the GPU to process multiple patches in parallel to exploit the massive parallelism of GPU devices.

\para{Row-wise Parallel Paeth Filter} Even within a single patch, it is necessary for GPU devices to exploit fine-grained parallelism to maximize performance. For the Paeth filter, we implement the decoder to process a set of pixels within a single row in parallel. Figure~\ref{fig:parallel}(b) presents the row-wise parallel processing of the Paeth filter. Because of both row-wise and column-wise data dependencies between three neighbor pixels (Section~\ref{sec:l3:design:algo}), the original Paeth filter should decode the pixels in a single patch sequentially from top-left to bottom-right. Our use of the custom Paeth filter makes parallelization easier as it inspects three top pixels in the preceding row for decoding, instead of the one left pixel and two top pixels (i.e., top, top-left) as in the original Paeth filter. Starting from the second row, pixels within a single row are decoded in parallel (i.e., each CUDA thread processes a single pixel in a row). Once the decoding for a row is completed, the decoding for the next row within the same patch begins. This process is repeated for all rows in the patch.

\para{Pixel-wise Parallel Base-Delta Decoding} Figure~\ref{fig:parallel}(c) shows the pixel-level parallel base-delta decoding in \name. The original execution takes the same order with the original Paeth filter--it reconstructs pixels from the first pixel to the last one sequentially in the patch. The \name decoder assigns a thread for base-delta decoding each pixel to exploit pixel-level parallelism within a row. Basically, each thread extracts the delta corresponding to the pixel and then adds the base value to reconstruct the original value.

\para{Overlapping Decoding and Gradient Computing on GPU} To execute both computing and decoding concurrently on the GPU, we allocate both processes to separate \emph{CUDA streams}. We prioritize the computing stream over the decoding stream to prevent the computing (processing the current batch) from being interfered with by the decoding (processing the subsequent batch).

%% file: contents/5-eval.tex
\begin{table}[t]
    \centering
    \caption{Detailed descriptions about models, datasets, and mini-batch sizes}
    \label{tab:batch}
    \begin{tabular}{c||cc}
        \hline
        \multirow{2}{*}{\textbf{\begin{tabular}[c]{@{}c@{}}Dataset (Resolution)\\ \# of train/val/test\end{tabular}}} & \multicolumn{1}{c|}{\multirow{2}{*}{\textbf{Model (Backbone)}}} & \multirow{2}{*}{\textbf{Batch size}} \\
        & \multicolumn{1}{c|}{} &  \\ \hline \hline
        \multirow{4}{*}{\begin{tabular}[c]{@{}c@{}}Cityscapes (1920$\times$1080)\\ 2975/500/1525\end{tabular}} & \multicolumn{1}{c|}{DDRNet23-slim} & 36 \\ \cline{2-3} 
        & \multicolumn{1}{c|}{DeepLabv3+ (MobileNetv2)} & 48 \\ \cline{2-3} 
        & \multicolumn{1}{c|}{MaskFormer (ResNet50)} & 32 \\ \cline{2-3} 
        & \multicolumn{1}{c|}{PointRend (SemanticFPN)} & 40 \\ \hline
        \multirow{3}{*}{\begin{tabular}[c]{@{}c@{}}KITTI (1024$\times$720)\\ 3519/3462/500\end{tabular}} & \multicolumn{1}{c|}{EgoNet} & 24 \\ \cline{2-3} 
        & \multicolumn{1}{c|}{PointPillars} & 48 \\ \cline{2-3} 
        & \multicolumn{1}{c|}{YOLOv5} & 128 \\ \hline
        DIV2K (2040$\times$1200) & \multicolumn{2}{c}{For compression ratio only} \\ \hline
        FFHQ (5760$\times$3840) & \multicolumn{2}{c}{For compression ratio only} \\ \hline
        RAISE-1K (4928$\times$3264) & \multicolumn{2}{c}{For compression ratio only} \\ \hline
    \end{tabular}
\end{table}

\section{Evaluation}
\label{sec:l3:eval}

\subsection{Methodology}
\label{sec:l3:eval:metho}

\para{Experimental Setup} We implement the decoder of \name by extending NVIDIA Data Loading Library (DALI)~\cite{dali:web} (version 1.6.0) by registering \name as a new image format. We initiate the training for each model by running DALI-integrated PyTorch (version 1.9.0)~\cite{pytorch:web}. For experiments, we use \texttt{p4d.24xlarge} AWS EC2 instance with 8$\times$NVIDIA A100 GPUs with 40GB HBM2 DRAM per each, 96 vCPUs on Intel Xeon Platinum 8275CL CPU, and 8$\times$1TB NVMe SSD.

\para{Models and Datasets} We compare \name with various lossless (PNG, JP2, BMP, LZ4, lossless WebP) and lossy (JPEG, lossy WebP) image formats and decoding method. Note that WebP support both lossless and lossy compression. For PNG, JP2, BMP, WebP, and JPEG, we use the implementation of the NVIDIA DALI framework~\cite{dali:web}. For LZ4 decoding we use NVIDIA \texttt{nvcomp} library~\cite{nvcomp:web}. We use seven object detection and semantic segmentation models (\texttt{DDR}: DDRNet23-slim~\cite{ddrnet2021hong}, \texttt{DL}: DeepLabv3+~\cite{deeplab2018chen}, \texttt{MF}: MaskFormer~\cite{maskformer2021cheng}, \texttt{PR}: PointRend~\cite{pointrend2020Kirillov}, \texttt{EN}: EgoNet~\cite{egonet2021li}, \texttt{PP}: PointPillars~\cite{pointpillar2019Lang}, and \texttt{YOLO}: YOLOv5~\cite{yolov5:web}) and two datasets (Cityscapes~\cite{cityscapes} and KITTI~\cite{kitti}) for throughput comparison. For the comparison of compression ratios of the lossless formats, we use three additional high-resolution datasets: DIV2K~\cite{div2k}, FFHQ~\cite{ffhq}, and RAISE-1K~\cite{raise1k}. Table~\ref{tab:batch} summarizes the model, dataset, and batch size used for training. The batch size is chosen to the maximum value for each model without causing an out-of-memory error. For the hyperparameters not shown in the table, we use the default values suggested in the papers or open-source implementations.

\para{Quality Factor of Lossy Image Formats} As we discuss in Section~\ref{sec:l3:back:acc}, a lossy image format may yield a lower metric performance than the lossless image format at the default configuration. For fair comparison, we tune the quality factor (Q-factor) of a lossy image format such that it achieves an equivalent metric performance to the lossless image format. Specifically, we set the Q-factor to be the minimum value while its accuracy falls within 1\% of that of the lossless image format. Table~\ref{tab:jpegqfac} reports the accuracy of L3, JPEG (Lossy), and WebP (Lossy) with the corresponding quality factors for the latter two. We use these values of Q-factor to compare the training throughput of JPEG and WebP (Lossy) with that of \name.

\begin{table}[t]
    \centering
    \caption{Test set accuracy of L3 (Lossless), JPEG (Lossy), and WebP (Lossy) and its quality factor}
    \label{tab:jpegqfac}
    \begin{tabular}{l||c||c|c}
        \hline
        \multicolumn{1}{c||}{\textbf{Model}} & \textbf{L3 acc.} & \textbf{JPEG acc. (Q-factor)} &  \textbf{WebP acc. (Q-factor)} \\ \hline \hline
        DDR & 0.729 & 0.730 (95) & 0.726 (95) \\ \hline
        DL & 0.801 & 0.801 (75) & 0.808 (75) \\ \hline
        MF & 0.685 & 0.686 (90) & 0.685 (95) \\ \hline
        PR & 0.725 & 0.721 (85) & 0.722 (85) \\ \hline
        EN & 0.323 & 0.323 (80) & 0.324 (80) \\ \hline
        PP & 0.611 & 0.613 (85) & 0.616 (95) \\ \hline
        YOLO & 0.633 & 0.635 (85) & 0.635 (85) \\ \hline
    \end{tabular}
\end{table}

\begin{table}[t]
    \centering
    \caption{Data compression ratio (compressed size/decompressed size). Lower is better.}
    \label{tab:compratio}
    \begin{tabular}{c||ccccc|cc}
        \hline
        \multicolumn{1}{l||}{\textbf{}} & \textbf{Cityscapes} & \textbf{KITTI} & \textbf{DIV2K} & \textbf{FFHQ} & \textbf{RAISE1K} & \textbf{Random} & \textbf{Black} \\ \hline \hline
        \textbf{Raw (MB)} & 5.93 & 2.24 & 7.01 & 63.28 & 46.02 & 5.93 & 5.93 \\ \hline
        \textbf{PNG} & 0.39$\times$ & 0.62$\times$ & 0.67$\times$ & 0.25$\times$ & 0.57$\times$ & 1.01$\times$ & 0.001$\times$ \\
        \textbf{JP2} & 0.38$\times$ & 0.57$\times$ & 0.59$\times$ & 0.28$\times$ & 0.56$\times$ & 1.09$\times$ & 0.005$\times$ \\
        \textbf{BMP} & 1.00$\times$ & 1.00$\times$ & 1.00$\times$ & 1.00$\times$ & 1.00$\times$ & 1.00$\times$ & 1.00$\times$ \\
        \textbf{LZ4} & 0.79$\times$ & 0.85$\times$ & 0.83$\times$ & 0.44$\times$ & 0.69$\times$ & 1.01$\times$ & 0.026$\times$ \\
        \textbf{WebP} & 0.29$\times$ & 0.55$\times$ & 0.49$\times$ & 0.15$\times$ & 0.37$\times$ & 1.00$\times$ & 0.000002$\times$ \\ \hline
        \textbf{\name} & 0.44$\times$ & 0.64$\times$ & 0.76$\times$ & 0.33$\times$ & 0.63$\times$ & 1.02$\times$ & 0.13$\times$ \\ \hline
    \end{tabular}
\end{table}

\subsection{Compression Ratio}
\label{sec:l3:eval:comp}

Table~\ref{tab:compratio} reports the compression ratio of the lossless formats, including \name, where lower is better. For this experiment, we use five image datasets, as well as two additional synthetic images. The random image (Random) selects a random value for each pixel in the range of 0 through 255. The black image (Black) sets all pixel values to zero. Except for the black image, the compression ratio of \name is worse than PNG and WebP (Lossless), falling within 6-9\% of the PNG and 9-30\% of the WebP (Lossless) format. However, we reiterate that the design objective of \name is not to maximize the compression ratio but to balance the DNN training pipeline with more balanced resource utilization. In particular, L3 eliminates the data preparation bottleneck by (i) maximizing decoding throughput while (ii) providing a good-enough compression to not cause data stalls on the \texttt{Load} stage.
\name's compression ratio is substantially worse in Black because \name does not incorporate a recurring patterns compression (e.g., run-length encoding, Huffman coding). However, we find that containing many trivially repeated patterns is not common in lossless, high-resolution images.

\begin{figure}[t]
    \centering
    \includegraphics[width=0.94\columnwidth]{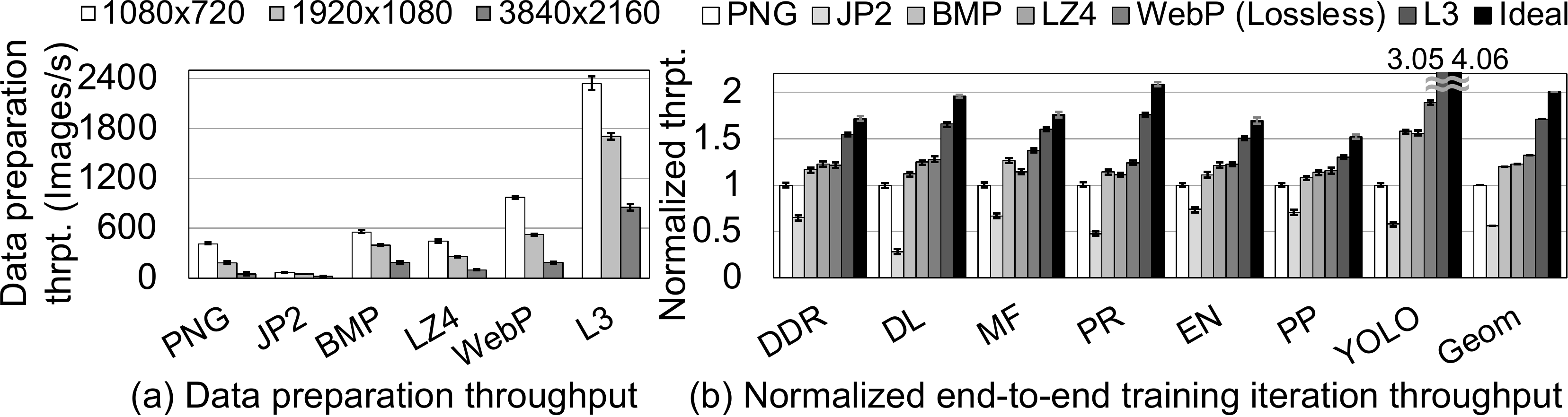}
    \caption{Throughput comparison with lossless-encoded datasets (a) Data preparation (\texttt{Load}+\texttt{Decode}) throughput (b) Normalized end-to-end training iteration throughput}
    \label{fig:lldecode}
\end{figure}

\subsection{Throughput Comparison with Lossless Decoders}
\label{sec:l3:eval:lossless}

\para{Data Preparation Throughput} Figure~\ref{fig:lldecode}(a) compares the data preparation throughput (\texttt{Load}+\texttt{Decode}) of \name and other lossless decoding formats with varying resolutions. We utilize the FHD Cityscapes dataset and its scaled-down/up versions using the Lanczos filter in Python Pillow package, which provides the highest quality for image down/upscaling~\cite{pil:web}. Compared to PNG, \name improves the data preparation throughput by 5.67$\times$, 9.29$\times$, and 15.71$\times$ for HD, FHD, and UHD images, respectively. Furthermore, \name outperforms WebP (Lossless), a state-of-the-art lossless image format, by a factor of 2.41$\times$, 3.25$\times$, and 4.51$\times$ for the same datasets. Thus, \name achieves substantially higher data preparation throughput than all the other lossless formats, regardless of the resolution.

\para{DNN Training Throughput} Figure~\ref{fig:lldecode}(b) presents the normalized end-to-end training throughput (iterations/sec) of different image formats on various models. The training throughput is normalized to PNG. \emph{Ideal} refers to the case when the data preparation stage is completely hidden by the \texttt{Compute} stage, thus yielding zero overhead. The choice of the image format has a significant impact on the training throughput as the data preparation time can potentially bottleneck the DNN training pipeline.

Overall, \name achieves the best end-to-end training throughput with a 1.71$\times$ geomean speedup compared to PNG. PNG and JP2 achieve lower throughput mainly because their complex algorithms running on CPU make them slow. The training throughput of BMP is limited by the I/O bandwidth (\texttt{Load}). LZ4 decoding utilizes GPUs, but its throughput is much lower than \name, achieving only 1.23$\times$ throughput speedup compared to PNG. WebP (Lossless) is the most recently proposed lossless image format, which achieves a 1.32$\times$ higher geomean training throughput than PNG. In contrast, \name achieves a 1.71$\times$ higher throughput than PNG, which outperforms WebP (Lossless) by a significant margin.

\subsection{Throughput Comparison with Lossy Decoders}
\label{sec:l3:eval:lossy}

\begin{figure*}[t]
    \centering
    \includegraphics[width=0.94\textwidth]{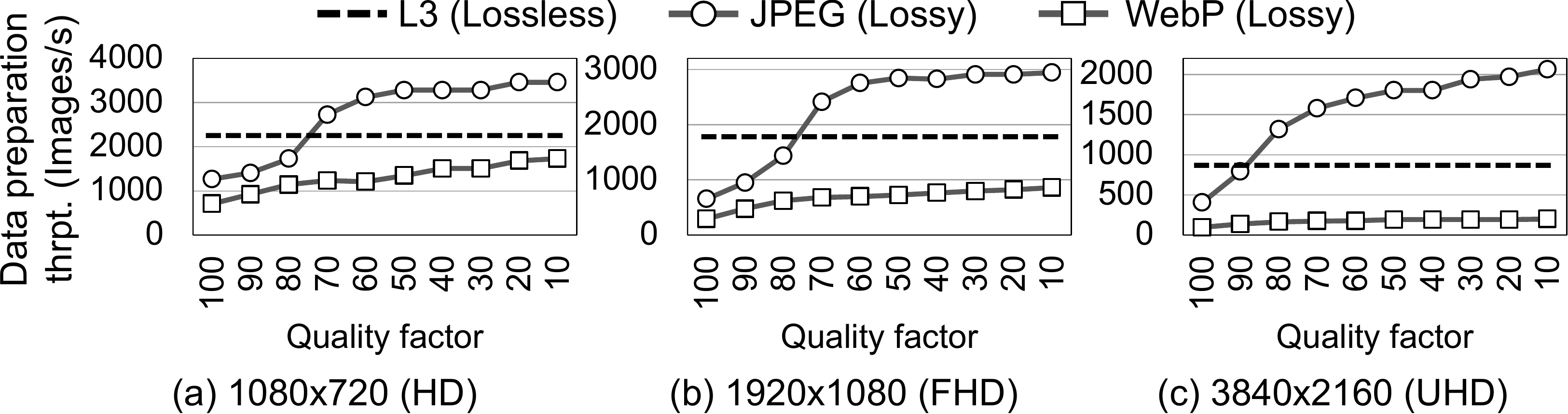}
    \caption{Data preparation throughput of JPEG (Lossy) and WebP (Lossy)-encoded Cityscapes datasets on various quality factors. (a) 1080$\times$720 (HD) (b) 1920$\times$1080 (FHD) (c) 3840$\times$2160 (UHD) resolution}
    \label{fig:jpegqfac}
\end{figure*}

\para{Impact of Q-Factor on Data Preparation Throughput} 
Figure~\ref{fig:jpegqfac} shows the trade-off between quality factor and data preparation throughput of JPEG (Lossy) and WebP (Lossy). We use the scaled-down/up versions of Cityscapes dataset and convert it to JPEG and WebP format. The black dotted line marks the data preparation throughput of \name (Lossless) for a given image resolution. WebP (Lossy) decoding is done on CPU, whereas JPEG decoding on the dedicated hardware JPEG accelerator on NVIDIA A100 GPU using \texttt{nvJPEG} library. Even with the lowest quality factor (i.e., highest decoding throughput), WebP (Lossy) has a lower decoding throughput than \name for all three image resolutions. This leads to a substantially lower end-to-end training throughput of WebP (Lossy) than \name.

\begin{figure*}[t]
    \centering
    \includegraphics[width=0.94\textwidth]{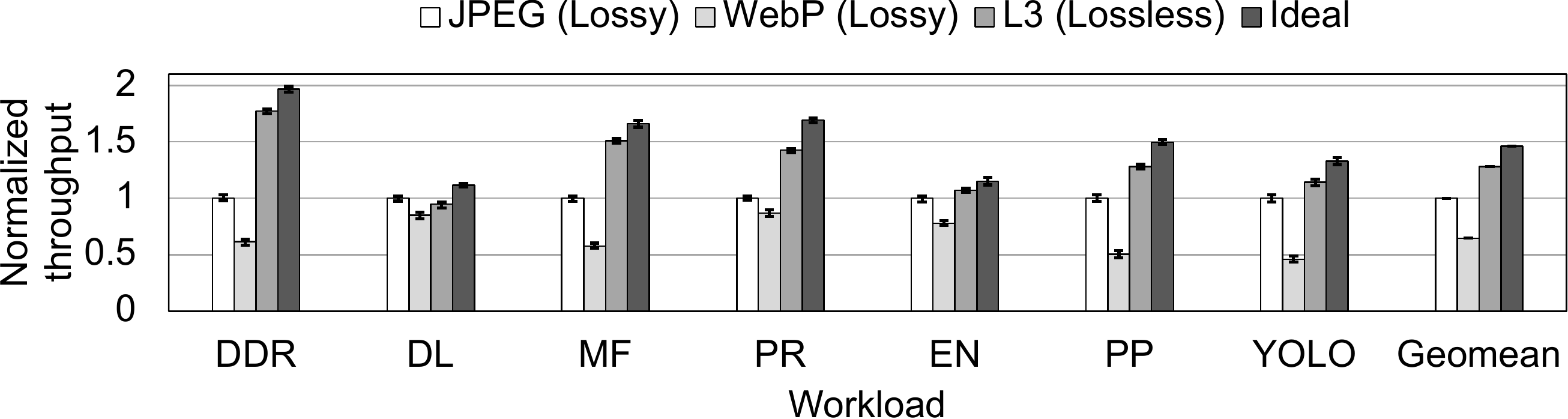}
    \caption{Normalized training iteration throughput with JPEG (Lossy), WebP (Lossy), and \name (Lossless)-encoded dataset}
    \label{fig:lossy}
\end{figure*}

In case of JPEG, there is a crossover point of the quality factor, where JPEG starts to outperform \name in terms of decoding throughput. As we increase the image resolution from HD to UHD, \name observes diminishing returns in throughput gains (in terms of pixels/sec) as the utilization of CUDA cores becomes high. JPEG performance scales a bit more gracefully to image resolution than \name as the entire decoder runs on a custom JPEG hardware accelerator which is available only on NVIDIA A100 GPU. However, \name can provide more robust performance on various data-parallel accelerators without requiring format-specific hardware support. In contrast, JPEG performance depends highly on the existence of custom hardware and enough CPU cores to sustain high throughput.

\para{DNN Training Throughput} Figure~\ref{fig:lossy} shows the end-to-end training iteration throughput for JPEG and WebP (Lossy) at equivalent test set accuracy. For comparable metric performance the quality factor of JPEG and WebP is set to the value in Table~\ref{tab:jpegqfac}). The throughput is normalized to that of JPEG. Overall, \name achieves up to 1.77$\times$ and 2.87$\times$ higher end-to-end training throughput than JPEG and WebP (Lossy), respectively. The geomean throughput gains are 1.25$\times$ for JPEG and 1.93$\times$ for WebP. Decreasing the quality factor may boost training throughput, but this comes with the cost of degrading the metric performance.

\begin{figure}[t]
    \centering
    \includegraphics[width=0.94\columnwidth]{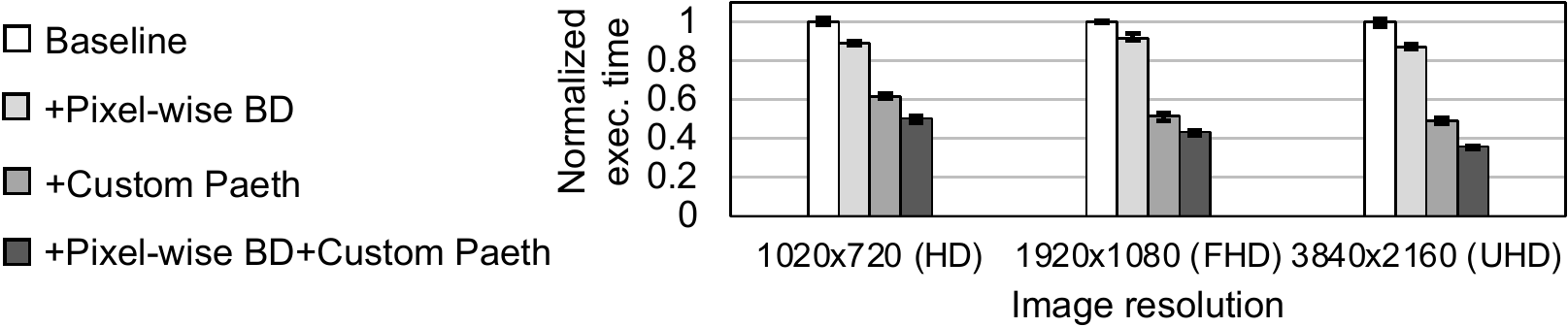}
    \caption{Normalized L3 decoder execution time on various image resolution (\texttt{Baseline}: Seq. original Paeth filter+Seq. base-delta decoding, \texttt{+Pixel-wise BD}: Baseline+pixel-wise parallel base-delta decoding, \texttt{+Custom Paeth}: Baseline+row-wise parallel custom Paeth filter, \texttt{+Pixel-wise BD+Custom Paeth}: Pixel-wise parallel base-delta decoding+Row-wise parallel custom Paeth filter)}
    \label{fig:module}
\end{figure}

\subsection{Ablation Study: Execution Time of L3 Decoder}
\label{sec:l3:eval:ablation}

L3 customizes the Paeth filter to make it more amenable to parallel execution by inspecting a different set of neighboring pixels (top-left, top, top-right), thus eliminating column-wise data dependency. Also, L3 unleashes pixel-level parallelism within the base-delta decoding step. Finally, L3 processes the decoding tasks in parallel with patch-level parallelism to exploit the massive parallelism of GPU devices. The patch-level parallelism reduces the decoding time by 76.8\%, 83.9\%, and 86.5\% for HD, FHD, and UHD resolutions, respectively. We set the state in which patch-level parallelism is applied as baseline.

Figure~\ref{fig:module} shows an ablation study to quantify the benefits of each component. The execution time is normalized by the baseline with sequential base-delta decoding and the original Paeth filter (Baseline). We use the FHD Cityscapes dataset and scale up/down the images with the Lanczos filter in Python Pillow package. The figure shows that pixel-wise parallel base-delta decoding (+Pixel-wise BD) and the custom Paeth filter (+Custom Paeth) reduces the decoding time by an average of 10.8\% and 46.0\%, respectively, over the baseline for the three image resolutions. With both optimizations applied (+Pixel-wise BD+Custom Paeth), the overall decoding time is reduced by 49.5\%, 56.7\%, and 59.1\% from the Baseline for HD, FHD, and UHD images, respectively.

%% file: contents/6-conclusion.tex
\section{Conclusion}
\label{sec:conc}

We propose \name, a new lightweight, lossless image format for high-resolution, high-throughput DNN training. The decoding algorithm of L3 is accelerated on data-parallel architectures such as GPUs. Thus, \name yields much faster decoding than the existing lossless image formats whose decoders are mostly running on the CPU. \name effectively eliminates data preparation bottlenecks in the DNN training pipeline. \name can be readily deployed on GPU without requiring any support for specialized hardware. \name can significantly reduce the end-to-end training time by providing higher throughput, higher accuracy, or both, compared to the existing lossless and lossy image formats at equivalent test set accuracy.

\quad

\para{Acknowledgements} This work was supported by SNU-SK Hynix Solution Research Center (S3RC), and National Research Foundation of Korea (NRF) grant funded by Korea government (MSIT) (NRF-2020R1A2C3010663). The source code is available at \url{https://github.com/SNU-ARC/L3.git}. Jae W. Lee is the corresponding author.

%% file: paper.bbl
\begin{thebibliography}{10}
\providecommand{\url}[1]{\texttt{#1}}
\providecommand{\urlprefix}{URL }
\providecommand{\doi}[1]{https://doi.org/#1}

\bibitem{tf2016martin}
Abadi, M., Barham, P., Chen, J., Chen, Z., Davis, A., Dean, J., Devin, M.,
  Ghemawat, S., Irving, G., Isard, M., Kudlur, M., Levenberg, J., Monga, R.,
  Moore, S., Murray, D.G., Steiner, B., Tucker, P., Vasudevan, V., Warden, P.,
  Wicke, M., Yu, Y., Zheng, X.: {TensorFlow}: A system for large-scale machine
  learning. In: Proceedings of the 12th {USENIX} Symposium on Operating Systems
  Design and Implementation. pp. 265--283. {USENIX} Association (November 2016)

\bibitem{div2k}
Agustsson, E., Timofte, R.: {NTIRE} 2017 challenge on single image
  super-resolution: Dataset and study. In: Proceedings of the IEEE Conference
  on Computer Vision and Pattern Recognition Workshops (July 2017)

\bibitem{deeplab2018chen}
Chen, L.C., Zhu, Y., Papandreou, G., Schroff, F., Adam, H.: Encoder-decoder
  with atrous separable convolution for semantic image segmentation. In:
  Proceedings of the European Conference on Computer Vision (September 2018)

\bibitem{mxnet2015chen}
Chen, T., Li, M., Li, Y., Lin, M., Wang, N., Wang, M., Xiao, T., Xu, B., Zhang,
  C., Zhang, Z.: {MXNet}: A flexible and efficient machine learning library for
  heterogeneous distributed systems. arXiv preprint arXiv:1512.01274  (2015)

\bibitem{tvm2018chen}
Chen, T., Moreau, T., Jiang, Z., Zheng, L., Yan, E., Shen, H., Cowan, M., Wang,
  L., Hu, Y., Ceze, L., Guestrin, C., Krishnamurthy, A.: {TVM}: An automated
  end-to-end optimizing compiler for deep learning. In: Proceedings of the 13th
  {USENIX} Symposium on Operating Systems Design and Implementation. pp.
  578--594. {USENIX} Association (October 2018)

\bibitem{Chen_2017_CVPR}
Chen, X., Ma, H., Wan, J., Li, B., Xia, T.: Multi-view {3D} object detection
  network for autonomous driving. In: Proceedings of the IEEE Conference on
  Computer Vision and Pattern Recognition (July 2017)

\bibitem{maskformer2021cheng}
Cheng, B., Schwing, A., Kirillov, A.: Per-pixel classification is not all you
  need for semantic segmentation. In: Proceedings of the Advances in Neural
  Information Processing Systems. vol.~34, pp. 17864--17875. Curran Associates,
  Inc. (December 2021)

\bibitem{cityscapes}
Cordts, M., Omran, M., Ramos, S., Rehfeld, T., Enzweiler, M., Benenson, R.,
  Franke, U., Roth, S., Schiele, B.: The cityscapes dataset for semantic urban
  scene understanding. In: Proceedings of the IEEE Conference on Computer
  Vision and Pattern Recognition (June 2016)

\bibitem{raise1k}
Dang-Nguyen, D.T., Pasquini, C., Conotter, V., Boato, G.: {RAISE}: A raw images
  dataset for digital image forensics. In: Proceedings of the 6th ACM
  Multimedia Systems Conference. p. 219–224. Association for Computing
  Machinery (March 2015)

\bibitem{imagenet2009jia}
Deng, J., Dong, W., Socher, R., jia Li, L., Li, K., Fei-fei, L.: {ImageNet}: A
  large-scale hierarchical image database. In: Proceedings of the IEEE
  Conference on Computer Vision and Pattern Recognition. IEEE (June 2009)

\bibitem{bd1991farrens}
Farrens, M., Park, A.: Dynamic base register caching: A technique for reducing
  address bus width. In: Proceedings of the 18th Annual International Symposium
  on Computer Architecture. p. 128–137. Association for Computing Machinery
  (May 1991)

\bibitem{all2017shunji}
Funasaka, S., Nakano, K., Ito, Y.: Adaptive loss-less data compression method
  optimized for {GPU} decompression. Concurrency and Computation: Practice and
  Experience  \textbf{29}(24),  e4283 (August 2017)

\bibitem{kitti}
Geiger, A., Lenz, P., Urtasun, R.: Are we ready for autonomous driving? {T}he
  {KITTI} vision benchmark suite. In: Proceedings of the IEEE Conference on
  Computer Vision and Pattern Recognition (June 2012)

\bibitem{ddrnet2021hong}
Hong, Y., Pan, H., Sun, W., Jia, Y.: Deep dual-resolution networks for
  real-time and accurate semantic segmentation of road scenes. arXiv preprint
  arXiv:2101.06085  (September 2021)

\bibitem{hou2019high}
Hou, L., Cheng, Y., Shazeer, N., Parmar, N., Li, Y., Korfiatis, P., Drucker,
  T.M., Blezek, D.J., Song, X.: High resolution medical image analysis with
  spatial partitioning. arXiv preprint arXiv:1909.03108  (September 2019)

\bibitem{gpip32019huang}
Huang, Y., Cheng, Y., Bapna, A., Firat, O., Chen, D., Chen, M., Lee, H., Ngiam,
  J., Le, Q.V., Wu, Y., Chen, z.: {GPipe}: Efficient training of giant neural
  networks using pipeline parallelism. In: Proceedings of the Advances in
  Neural Information Processing Systems. vol.~32. Curran Associates, Inc.
  (December 2019)

\bibitem{ffhq}
Karras, T., Laine, S., Aila, T.: A style-based generator architecture for
  generative adversarial networks. In: Proceedings of the IEEE/CVF Conference
  on Computer Vision and Pattern Recognition (June 2019)

\bibitem{pointrend2020Kirillov}
Kirillov, A., Wu, Y., He, K., Girshick, R.: {PointRend}: Image segmentation as
  rendering. In: Proceedings of the IEEE/CVF Conference on Computer Vision and
  Pattern Recognition (June 2020)

\bibitem{cifar}
Krizhevsky, A., Hinton, G.: Learning multiple layers of features from tiny
  images. In: Technical report. Citeseer (April 2009)

\bibitem{pointpillar2019Lang}
Lang, A.H., Vora, S., Caesar, H., Zhou, L., Yang, J., Beijbom, O.:
  {PointPillars}: Fast encoders for object detection from point clouds. In:
  Proceedings of the IEEE/CVF Conference on Computer Vision and Pattern
  Recognition (June 2019)

\bibitem{egonet2021li}
Li, S., Yan, Z., Li, H., Cheng, K.T.: Exploring intermediate representation for
  monocular vehicle pose estimation. In: Proceedings of the IEEE/CVF Conference
  on Computer Vision and Pattern Recognition. pp. 1873--1883 (June 2021)

\bibitem{swinir}
Liang, J., Cao, J., Sun, G., Zhang, K., Van~Gool, L., Timofte, R.: {SwinIR}:
  Image restoration using swin transformer. In: Proceedings of the IEEE/CVF
  International Conference on Computer Vision Workshops. pp. 1833--1844
  (October 2021)

\bibitem{edsr}
Lim, B., Son, S., Kim, H., Nah, S., Mu~Lee, K.: Enhanced deep residual networks
  for single image super-resolution. In: Proceedings of the IEEE Conference on
  Computer Vision and Pattern Recognition Workshops. pp. 136--144 (July 2017)

\bibitem{rammer2020ma}
Ma, L., Xie, Z., Yang, Z., Xue, J., Miao, Y., Cui, W., Hu, W., Yang, F., Zhang,
  L., Zhou, L.: Rammer: Enabling holistic deep learning compiler optimizations
  with {rTasks}. In: Proceedings of the 14th {USENIX} Symposium on Operating
  Systems Design and Implementation. pp. 881--897. {USENIX} Association
  (November 2020)

\bibitem{dragon2018markthub}
Markthub, P., Belviranli, M.E., Lee, S., Vetter, J.S., Matsuoka, S.: {DRAGON}:
  Breaking {GPU} memory capacity limits with direct {NVM} access. In:
  Proceedings of the International Conference for High Performance Computing,
  Networking, Storage, and Analysis. pp. 32:1--32:13. IEEE (November 2018)

\bibitem{Menze_2015_CVPR}
Menze, M., Geiger, A.: Object scene flow for autonomous vehicles. In:
  Proceedings of the IEEE Conference on Computer Vision and Pattern Recognition
  (June 2015)

\bibitem{mohan2020analyzing}
Mohan, J., Phanishayee, A., Raniwala, A., Chidambaram, V.: Analyzing and
  mitigating data stalls in {DNN} training. Proceedings of the VLDB Endowment
  \textbf{14}(5),  771–784 (Jan 2021)

\bibitem{tfdata2021murray}
Murray, D.G., Simsa, J., Klimovic, A., Indyk, I.: tf.data: A machine learning
  data processing framework. Proceedings of the VLDB Endowment
  \textbf{14}(12),  2945–2958 (July 2021)

\bibitem{pipedream2019deepak}
Narayanan, D., Harlap, A., Phanishayee, A., Seshadri, V., Devanur, N.R.,
  Ganger, G.R., Gibbons, P.B., Zaharia, M.: {PipeDream}: Generalized pipeline
  parallelism for {DNN} training. In: Proceedings of the 27th ACM Symposium on
  Operating Systems Principles. p. 1–15. Association for Computing Machinery
  (October 2019)

\bibitem{svhn}
Netzer, Y., Wang, T., Coates, A., Bissacco, A., Wu, B., Ng, A.Y.: Reading
  digits in natural images with unsupervised feature learning. In: Proceedings
  of the NIPS Workshop on Deep Learning and Unsupervised Feature Learning
  (December 2011)

\bibitem{a100hwacc}
NVIDIA: {NVIDIA} {A100} tensor core {GPU} architecture.
  \url{https://images.nvidia.com/aem-dam/en-zz/Solutions/data-center/nvidia-ampere-architecture-whitepaper.pdf}
  (2020)

\bibitem{nvcomp:web}
NVIDIA: nvcomp: A library for fast lossless compression/decompression on the
  {GPU}. \url{https://github.com/NVIDIA/nvcomp} (2021)

\bibitem{dali:web}
NVIDIA: The {NVIDIA} data loading library ({DALI}).
  \url{https://github.com/NVIDIA/DALI} (2021)

\bibitem{nvjpeg:web}
NVIDIA: nv{JPEG} libraries: {GPU}-accelerated {JPEG} decoder, encoder and
  transcoder. \url{https://developer.nvidia.com/nvjpeg} (2021)

\bibitem{nvjpeg2k:web}
NVIDIA: nv{JPEG}2000 libraries. \url{https://docs.nvidia.com/cuda/nvjpeg2000}
  (2021)

\bibitem{culzss2011}
Ozsoy, A., Swany, M.: {CULZSS}: {LZSS} lossless data compression on {CUDA}. In:
  Proceedings of the 2011 IEEE International Conference on Cluster Computing.
  pp. 403--411 (September 2011)

\bibitem{paeth1991}
Paeth, A.W.: {II}.9 - {I}mage file compression made easy. In: Graphics Gems II,
  pp. 93--100. Morgan Kaufmann (1991)

\bibitem{trainbox2020park}
Park, P., Jeong, H., Kim, J.: {TrainBox}: An extreme-scale neural network
  training server architecture by systematically balancing operations. In:
  Proceedings of the 2020 53rd Annual IEEE/ACM International Symposium on
  Microarchitecture. pp. 825--838 (October 2020)

\bibitem{bzipaccel}
Patel, R.A., Zhang, Y., Mak, J., Davidson, A., Owens, J.D.: Parallel lossless
  data compression on the {GPU}. In: Proceedings of the 2012 Innovative
  Parallel Computing. pp.~1--9 (May 2012)

\bibitem{bdi2012gennady}
Pekhimenko, G., Seshadri, V., Mutlu, O., Gibbons, P.B., Kozuch, M.A., Mowry,
  T.C.: Base-delta-immediate compression: Practical data compression for
  on-chip caches. In: Proceedings of the 21st International Conference on
  Parallel Architectures and Compilation Techniques. p. 377–388. Association
  for Computing Machinery (September 2012)

\bibitem{dlt2019}
Peng, Y., Zhu, Y., Chen, Y., Bao, Y., Yi, B., Lan, C., Wu, C., Guo, C.: A
  generic communication scheduler for distributed {DNN} training acceleration.
  In: Proceedings of the 27th ACM Symposium on Operating Systems Principles. p.
  16–29. Association for Computing Machinery (October 2019)

\bibitem{pil:web}
Pillow: Python pillow filters.
  \url{https://pillow.readthedocs.io/en/stable/handbook/concepts.html\#filters}
  (2021)

\bibitem{pytorch:web}
PyTorch: {PyTorch}. \url{https://pytorch.org} (2021)

\bibitem{brain2020}
Rebsamen, M., Suter, Y., Wiest, R., Reyes, M., Rummel, C.: Brain morphometry
  estimation: From hours to seconds using deep learning. Frontiers in Neurology
   \textbf{11}, ~244 (April 2020)

\bibitem{deam}
Ren, C., He, X., Wang, C., Zhao, Z.: Adaptive consistency prior based deep
  network for image denoising. In: Proceedings of the IEEE/CVF Conference on
  Computer Vision and Pattern Recognition. pp. 8596--8606 (June 2021)

\bibitem{vdnn2016rhu}
Rhu, M., Gimelshein, N., Clemons, J., Zulfiqar, A., Keckler, S.W.: {vDNN}:
  Virtualized deep neural networks for scalable, memory-efficient neural
  network design. In: Proceedings of the 49th Annual IEEE/ACM International
  Symposium on Microarchitecture. pp. 18:1--18:13. IEEE (October 2016)

\bibitem{cano2021}
Sarangi, S., Baas, B.: Canonical huffman decoder on fine-grain many-core
  processor arrays. In: Proceedings of the 2021 26th Asia and South Pacific
  Design Automation Conference. pp. 512--517 (January 2021)

\bibitem{gompresso2016Sitaridi}
Sitaridi, E., Mueller, R., Kaldewey, T., Lohman, G., Ross, K.A.:
  Massively-parallel lossless data decompression. In: Procceedings of the 2016
  45th International Conference on Parallel Processing. pp. 242--247 (August
  2016)

\bibitem{yolov5:web}
Ultralytics: Yolov5. \url{https://github.com/ultralytics/yolov5/} (2021)

\bibitem{super2018wang}
Wang, L., Ye, J., Zhao, Y., Wu, W., Li, A., Song, S.L., Xu, Z., Kraska, T.:
  {SuperNeurons}: Dynamic {GPU} memory management for training deep neural
  networks. In: Proceedings of the 23rd ACM SIGPLAN Symposium on Principles and
  Practice of Parallel Programming. pp. 41--53. ACM (February 2018)

\bibitem{diesel2020wang}
Wang, L., Ye, S., Yang, B., Lu, Y., Zhang, H., Yan, S., Luo, Q.: {DIESEL}: A
  dataset-based distributed storage and caching system for large-scale deep
  learning training. In: Proceedings of the 49th International Conference on
  Parallel Processing. Association for Computing Machinery (August 2020)

\bibitem{esrgan}
Wang, X., Yu, K., Wu, S., Gu, J., Liu, Y., Dong, C., Qiao, Y., Change~Loy, C.:
  {ESRGAN}: Enhanced super-resolution generative adversarial networks. In:
  Proceedings of the European Conference on Computer Vision Workshops
  (September 2018)

\bibitem{uformer}
Wang, Z., Cun, X., Bao, J., Zhou, W., Liu, J., Li, H.: Uformer: A general
  u-shaped transformer for image restoration. In: Proceedings of the IEEE/CVF
  Conference on Computer Vision and Pattern Recognition. pp. 17683--17693 (June
  2022)

\bibitem{massive2018}
Wei\ss{}enberger, A., Schmidt, B.: Massively parallel huffman decoding on
  {GPUs}. In: Proceedings of the 47th International Conference on Parallel
  Processing. Association for Computing Machinery (August 2018)

\bibitem{eunet}
Xu, L., Zhang, J., Cheng, X., Zhang, F., Wei, X., Ren, J.: Efficient deep image
  denoising via class specific convolution. Proceedings of the AAAI Conference
  on Artificial Intelligence  \textbf{35}(4),  3039--3046 (May 2021)

\bibitem{huffman2020yamamoto}
Yamamoto, N., Nakano, K., Ito, Y., Takafuji, D., Kasagi, A., Tabaru, T.:
  Huffman coding with gap arrays for {GPU} acceleration. In: Proceedings of the
  49th International Conference on Parallel Processing. Association for
  Computing Machinery (August 2020)

\bibitem{medi2020}
Zhou, S., Nie, D., Adeli, E., Yin, J., Lian, J., Shen, D.: High-resolution
  encoder–decoder networks for low-contrast medical image segmentation. IEEE
  Transactions on Image Processing  \textbf{29},  461--475 (2020)

\bibitem{Zhou_2018_CVPR}
Zhou, Y., Tuzel, O.: {VoxelNet}: End-to-end learning for point cloud based {3D}
  object detection. In: Proceedings of the IEEE Conference on Computer Vision
  and Pattern Recognition (June 2018)

\end{thebibliography}
